\documentclass{article}
\usepackage[utf8]{inputenc}
\usepackage[T1]{fontenc}
\usepackage[english]{babel}
\usepackage{amsmath}
\usepackage{amssymb,amsfonts,textcomp}
\usepackage{array}
\usepackage{supertabular}
\usepackage{hhline}
\usepackage[pdftex]{graphicx}
\makeatletter
\newcommand\arraybslash{\let\\\@arraycr}
\makeatother
\begin{document}
\title{The Effects of Hyperparameters on SGD Training of Neural Networks}
\author{Thomas M. Breuel \\ Google, Inc. \\ {\tt tmb@google.com}}
\date{}
\maketitle

\begin{abstract}
The performance of neural network classifiers is determined by a number
of hyperparameters, including learning rate, batch size, and depth. A
number of attempts have been made to explore these parameters in the
literature, and at times, to develop methods for optimizing them.
However, exploration of parameter spaces has often been limited. In
this note, I report the results of large scale experiments exploring
these different parameters and their interactions.
\end{abstract}

\section{Datasets and Libraries}

All experiments reported here were carried out using the Torch library
\cite{collobert2002torch} and CUDA (some of the experiments have been reproduced on a
smaller scale with other libraries). The dataset for all the
experiments is MNIST \cite{lecun1998mnist,keysers2007comparison}.

Characters were deskewed prior to all experiments. Deskewing
significantly reduces error rates in nearest neighbor classifiers. Skew
corresponds to a simple one-parameter family of linear transformations
in feature space and causes decision regions to become highly
anisotropic. Without deskewing, differences in performance between
different architectures might primarily reduce to their ability to
“learn deskewing”. With deskewing, MNIST character classification
become more of an instance of a typical classification problem. Prior
results on classifying deskewed MNIST data both with neural networks
and with other methods are shown in the table below.

\begin{table}[tp]
\begin{flushleft}
\tablehead{}
\begin{tabular}{|p{1.5in}|r|p{1in}|p{1.5in}|}
\hline
 Method &
 Test Error &
 Preprocessing &
 Reference\\\hline
Reduced Set SVM deg 5 polynomial &
1 &
 deskewing &
 LeCun et al. 1998\\\hline
 SVM deg 4 polynomial &
1.1 &
 deskewing &
 LeCun et al. 1998\\\hline
 K-nearest-neighbors, L3 &
1.22 &
 deskewing, noise removal, blurring, 2 pixel shift &
 Kenneth Wilder, U. Chicago\\\hline
 K-nearest-neighbors, L3 &
1.33 &
 deskewing, noise removal, blurring, 1 pixel shift &
 Kenneth Wilder, U. Chicago\\\hline
 2-layer NN, 300 HU &
1.6 &
 deskewing &
 LeCun et al. 1998\\\hline
\end{tabular}
\end{flushleft}
\caption{\label{tab-mnist-results}
Other previously reported results on the MNIST database.
}
\end{table}

\section{Logistic vs Softmax Outputs}

Multi-Layer Perceptrons (MLPs)
used for classification usually attempt to approximate posterior
probabilities and use those as their discriminant function. Two common
approaches to this are the use of least square regression with logistic
output units trained with a least square error measure (“logistic 
outputs”) and a softmax output layer (“softmax outputs”). 
In the limit of infinite amounts of training data, both
approaches converge to true posterior probability estimates. Softmax
output layers have the property that they are guaranteed to produce a
normalized posterior probability distribution across all classes, while
least square regression with logistic output units generates
independent probability estimates for each class membership without any
guarantees that these probabilities sum up to one.

Softmax is often preferred, although there is no obvious theoretical
reason why it should yield better discriminant functions or lower
classification error for finite training sets. In OCR and speech
recognition, some practitioners have observed that logistic outputs
yield better posterior probability estimates and better results when
combined with probabilistic language models. In addition, when the sum
of the posterior probability estimates derived from logistic outputs
differs significantly from unity, that is a strong indication that the
input lies outside the training set and should be rejected.

\begin{figure}[tp]
\includegraphics[height=3in]{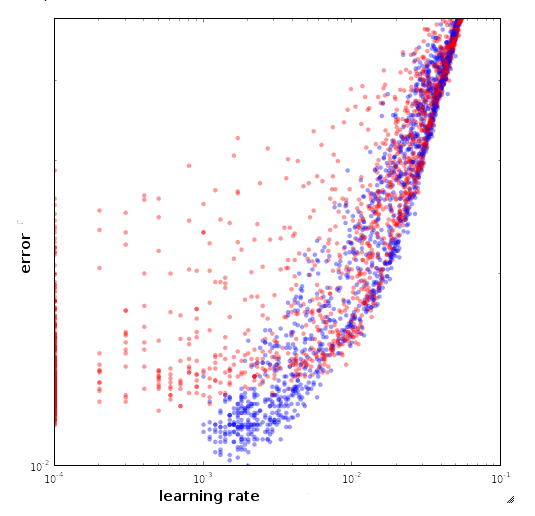}
\caption{\label{mlpsoftmax}
Training and test error for MLPs with logistic outputs (blue)
and softmax output (red). Note that softmax outputs achieve down to
zero percent training error (assigned to an error of 1e-4) but
logistic outputs give overall better performance on new training
samples.
}
\end{figure}

Figure~\ref{mlpsoftmax} shows a scatterplot of test vs training error for a large
number of MLPs with one hidden layer at different learning rates,
different number of hidden units, and different batch sizes. Such
scatterplots show what error rates are achievable by the different
architectures, hyperparameter choices, initializations, and order of
sample presentations.  The lowest points in the vertical direction
indicate the lowest test set error achievable by the architecture in
this set of experiments. The scatterplot shows that logistic outputs
achieve test set error rates of about 1.0\% vs 1.1\% for softmax
outputs. At the same time, logistic outputs never achieve zero percent
training set error, while softmax outputs frequently do.

\begin{figure}[tp]
\includegraphics[width=\textwidth]{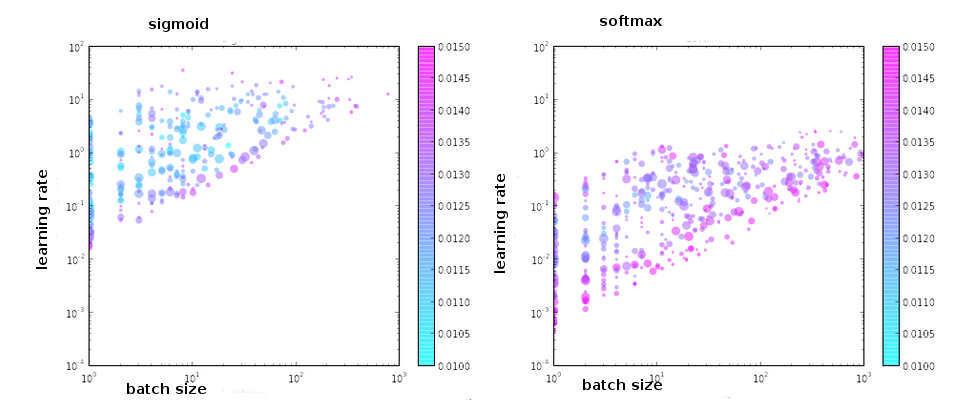}
\caption{\label{lrbatch}
Learning rate and batch size effects depending on output layer
type. This is a scatterplot for all networks that yield test set error
rates of less than 1.5\%, with color indicating the test set error.
Softmax outputs yield the best results for a learning rate that is
about an order of magnitude smaller than logistic outputs.
}
\end{figure}

In order to ascertain that the difference in test set error between the
two architectures is due to the architectures themselves, it is
important to ensure that the space of hyperparameters (learning rates,
batch sizes, number of hidden units) has been explored sufficiently.
Note that, as Figure~\ref{lrbatch} shows, the in order to yield low error rates,
softmax outputs require learning rates that are about an order of
magnitude lower than logistic outputs.

\begin{figure}[tp]
\includegraphics[height=3in]{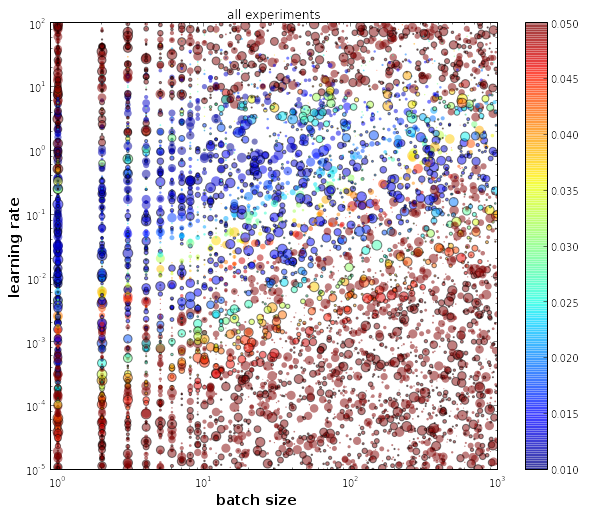}
\caption{\label{parange}
The complete range of parameters explored in the MLP
experiments reported in this section. In this plot, color indicates
error rate; circle size indicates the number of hidden units; a circle
with a border represents softmax outputs.
}
\end{figure}

Figure~\ref{parange} demonstrates that the range of parameters (learning rates,
batch sizes) has been explored fully; above learning rates of 1e1,
both softmax and logistic output models diverge, and at small learning
rates, both fail to learn in a reasonable amount of time.

The fact that logistic outputs yield 10\% lower relative error rates on
such a simple and widely studied problem and architecture compared to
softmax outputs does not prove that “logistic outputs are better than
softmax outputs”, but it suggests that it is worth testing both
logistic and softmax outputs on any particular problem to see which one
yields lower test set error.

\section{Batch size Effects}

Optimizing the weights of a neural network can be carried out by
stochastic gradient descent (updating after each sample) or by full
gradient descent (computing a gradient on the parameters from the
entire training set). In between those two extremes is batched gradient
descent. In batched gradient descent, we update the parameters of the
MLP to reduce the error for a small sample (“batch”) of training
samples, typically consisting of between 10 and 1000 samples.

Computationally, using batches instead of individual training samples
allows for greater parallelism; each layer of an MLP computes
effectively a function like:

$$ y = \sigma(M\cdot x) $$

For simple SGD, $x$ is some d{}-dimensional vector, but for batched gradient descent,
$x$ is a $d\times b$ dimensional matrix, where
$b$ is the batch size. The matrix multiplication
$M\cdot x$ can be evaluated much more efficiently and in parallel than evaluating
$b$ individual matrix-vector products in single-sample updates. In fact,
if we use $b$ processors, we can compute updates for
$b$ samples in roughly the same time as we would otherwise use for a
single sample in SGD.

When using batch training, a common convention is to rescale the
learning rate $\lambda \rightarrow \frac{\lambda}{b}$
This means that as we increase the batch size $b$,
we need to scale up the learning rate proportionately. This is the
convention we use in these experiments.

\begin{figure}[tp]
\includegraphics[height=3in]{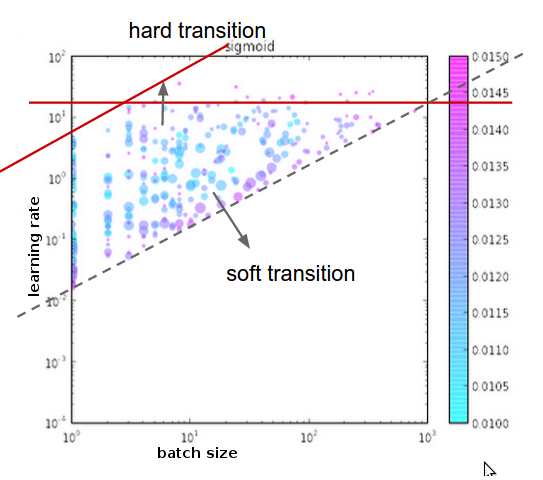}
\caption{\label{bslr}
Batch size and learning rate vs. error rate (logistic outputs).
See the text for an explanation.
}
\end{figure}

Figure~\ref{bslr} shows a scatterplot of trained networks with good test set
error vs. batch size and learning rate. There are three apparent
limits on performance:

\begin{enumerate}
\item At the lower end, we have a soft transition from well performing
networks to poorly performing networks. The explanation of this is that
at low learning rates, the network learns too slowly to reach low test
set errors within the limited number of training steps. The reason this
soft transition slopes upwards is due to the use of normalized learning
rates. Without learning rate normalization, this line would remain
horizontal and independent of batch size.
\item At a batch size of 1, there is a maximum learning rate; beyond
that learning rate, the stochastic gradient descent optimization
diverges. Without batch size normalization of learning rate, this upper
limit would exist independent of batch size; due to batch size
normalization, this line of divergence slopes upwards, parallel to
the soft lower bound on learning rates.
\item There is a third, unexpected, constant limit on the
batch-normalized learning rate.
\end{enumerate}

The original single sample learning rate determines the speed of
convergence of the stochastic gradient descent algorithm; within its
region of convergence, halving the learning rate approximately doubles
the time needed to reach a given test set error, since each gradient
update represents simply a step towards the minimum along some path.
This reasoning also applies for batch updates. The constant upper limit
on the batch-normalized learning rate corresponds to a single sample
learning rate that decreases proportionally to batch size.

The consequence is that, as long as the maximum usable batch normalized
learning rate increases proportionally with batch size, we benefit from
parallelization in terms of overall speedup of learning. Once we enter
the regime where the upper limit of batch normalized learning rates is
independent of batch size, further parallelization does not speed up
training.

Without further experimentation, we can only guess at the source of the
upper limit on the batch-normalized learning rate. As a simplified
model, assume that the limit on the learning rate for single sample
updates is due to some subset of the input vectors (e.g., vectors that
generate particularly large gradients). For batch sizes that contain,
on average, only one of those input vectors, we can continue to use the
original learning rate, but once we use a batch size that contains, on
average, two of those vectors, we have to cut the learning rate in half
in order to keep the magnitude of the update within the range that
allows convergence. (In practice, we are not necessarily looking at
individual samples but subspaces of the input.) This analysis suggests
possible strategies for improving training performance with large batch
sizes that will be explored elsewhere.

\begin{figure}[tp]
\includegraphics[width=\textwidth]{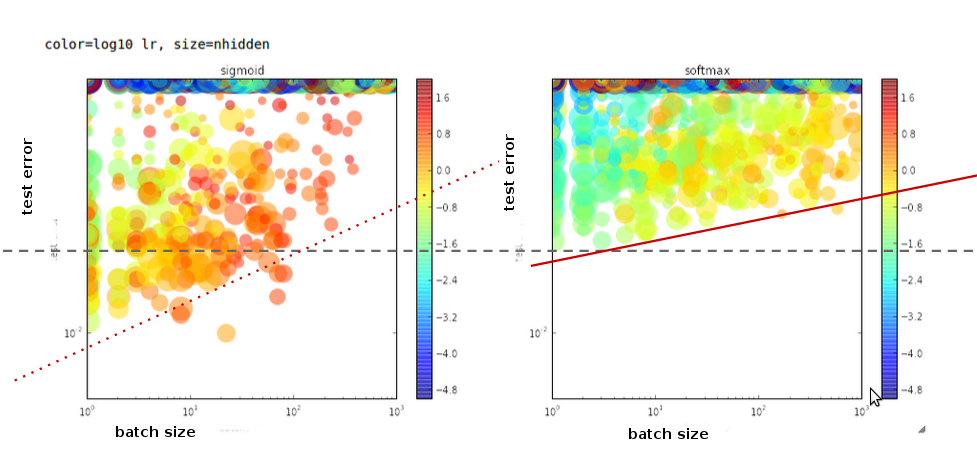}
\caption{\label{bste}
Test error by batch size for logistic and softmax outputs. Note
that logistic output units achieve the lowest error. Also note that for
larger batch sizes, the advantage of logistic output units over softmax
output units disappears.
}
\end{figure}

Regardless of the speed of optimization, we can also ask the question of
how the test set error of the final network depends on batch size. This
is shown in Figure~\ref{bste}. We see that increasing batch size generally
results in worse test set errors for both logistic outputs and softmax
outputs. The dependence is somewhat stronger for logistic outputs. In
addition, logistic outputs appear to yield networks with a higher
variability. Note that the differences in error rates in this plot are
much smaller than the differences in error rates found in the 
previous learning rate plots; this plot makes small but significant
differences among the very best models visible.

\begin{figure}[tp]
\includegraphics[height=3in]{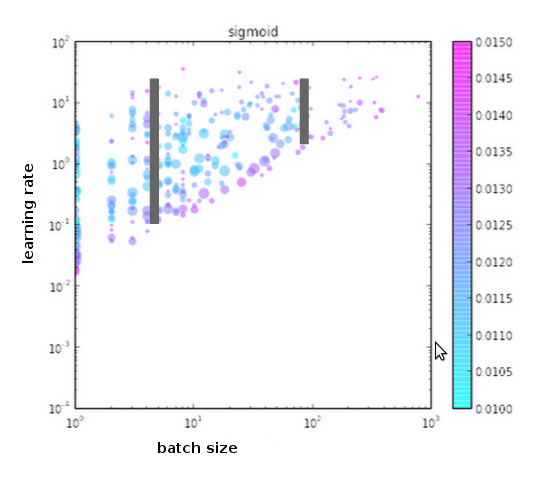}
\caption{\label{hypnar}
At large batch sizes, the range of hyperparameter resulting in
good test set errors is considerably smaller than at small batch sizes.
}
\end{figure}

The observations on the relationship between batch sizes and learning
rates above also have implications for hyperparameter optimization. In
particular, the hyperparameter search at large batch sizes becomes
harder because the range of learning rates that yield good networks is
considerably smaller than it is at small batch sizes (Figure~\ref{hypnar}). 
Batch sizes that are too large therefore not only waste computational
resources through parallelism that does not result in a speedup for
learning, they may actually make the hyperparameter search harder.

\section{Convolutional Layers}

The above results were all obtained for non-convolutional networks with
a single hidden layer. How do they generalize to convolutional
networks? There are, of course, many different kinds of convolutional
architectures we could investigate. The simplest architecture places a
single convolutional layer at the input of the network.

\begin{figure}[tp]
\includegraphics[height=3in]{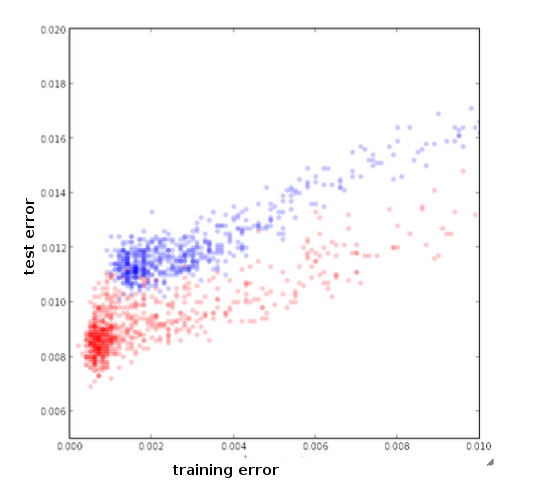}
\caption{\label{convscatter}
Training error vs test set error for non-convolutional (blue)
and convolutional (red) networks.
}
\end{figure}

Not surprisingly, adding a convolutional layer results in significantly
lower test set error (0.69\% test set error) compared to
non-convolutional networks (1\% test set error), as seen in Figure~\ref{convscatter}.

\begin{figure}[tp]
\includegraphics[height=3in]{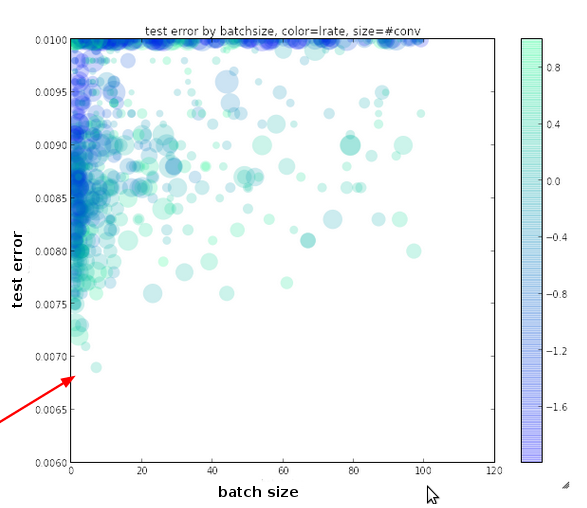}
\caption{\label{bsconv}
Test set error by batch size for convolutional networks.
Output units are softmax, and the plot shows the best networks
across all learning rates and number of convolutional
units.  Note the significant increase in best achievable error rates with
increasing batch size.
}
\end{figure}

For convolutional networks, we also observe that small batch sizes yield
the best test set errors. In fact, large batch sizes never reach
comparably low error rates (Figure~\ref{bsconv}).

\section{ReLU Units (Not Convolutional)}

Another popular architectural choice is to replace the sigmoidal
nonlinearity in hidden layers with rectifying linear units (ReLU). To
explore the effects of this on the results, networks were trained with
all four combinations of sigmoid/ReLU hidden units and softmax/sigmoid
output units. The scatterplot of results is shown in Figure 9.

\begin{figure}[tp]
\hbox{
\includegraphics[height=3in]{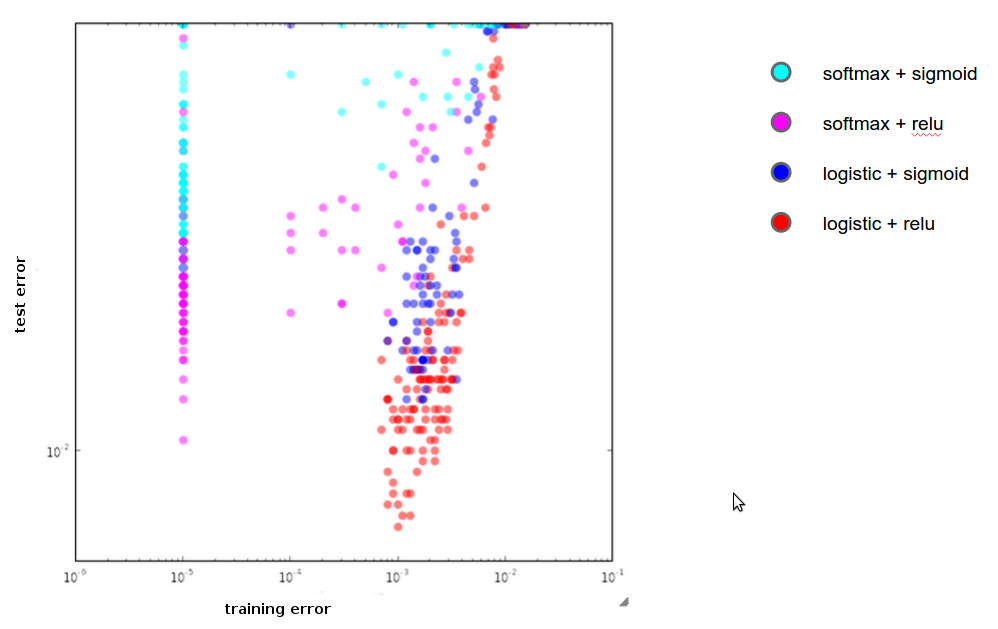}
}
\caption{\label{sigrel}
A comparison of sigmoid and ReLU units at the hidden layer,
combined with softmax and logistic outputs.
}
\end{figure}

We see in Figure~\ref{sigrel}
that softmax outputs achieve zero percent test set error for both
kinds of hidden layer nonlinearities. Furthermore, ReLU hidden units
outperform sigmoidal hidden units for either kind of output layer. The
overall best performing combination was logistic output units with ReLU
hidden units, resulting in a test set error of 0.92\%.

\begin{figure}[tp]
\includegraphics[height=3in]{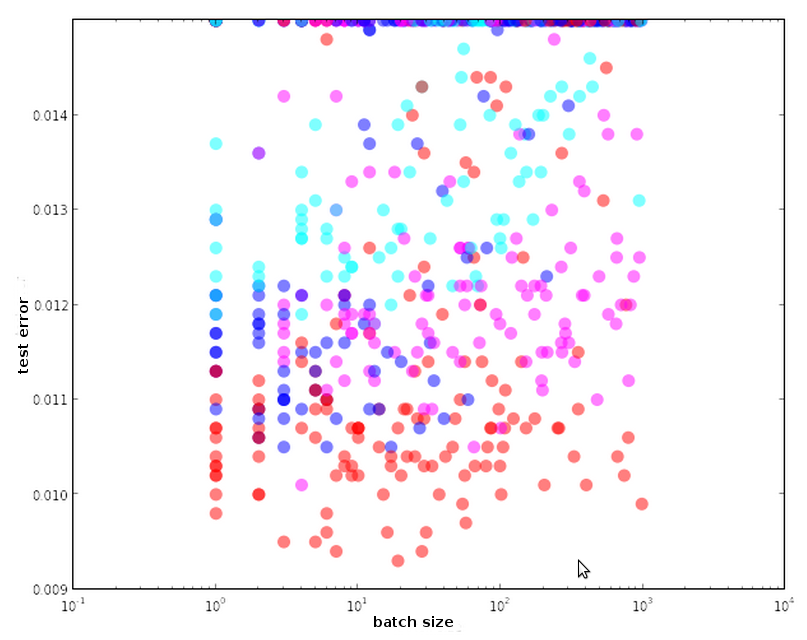}
\caption{\label{tsbatch}
Dependence of test set error on batch size. We see that for
sigmoidal hidden units (blue, cyan), large batch sizes perform
considerably worse than small batch sizes. For ReLU hidden units (red,
magenta), the dependence is considerably weaker, although smaller
batch sizes still seem to have a slight advantage.
}
\end{figure}

\begin{figure}[tp]
\includegraphics[height=3in]{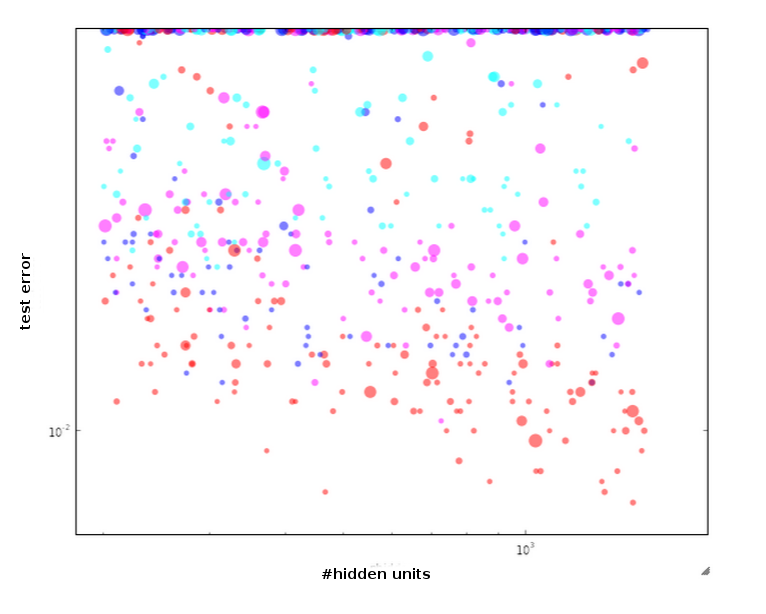}
\caption{\label{tshidden}
Dependence of test set error on the number of hidden units.
For sigmoidal hidden units (blue, cyan), there is little improvement of
test set error with increasing numbers of hidden units. For ReLU hidden
units, there is a strong improvement in test set error with the number
of hidden units. The maximum number of hidden units tested was 2000,
although it looks like larger numbers of hidden units might result in
even better performance.
}
\end{figure}

\begin{figure}[tp]
\includegraphics[height=3in]{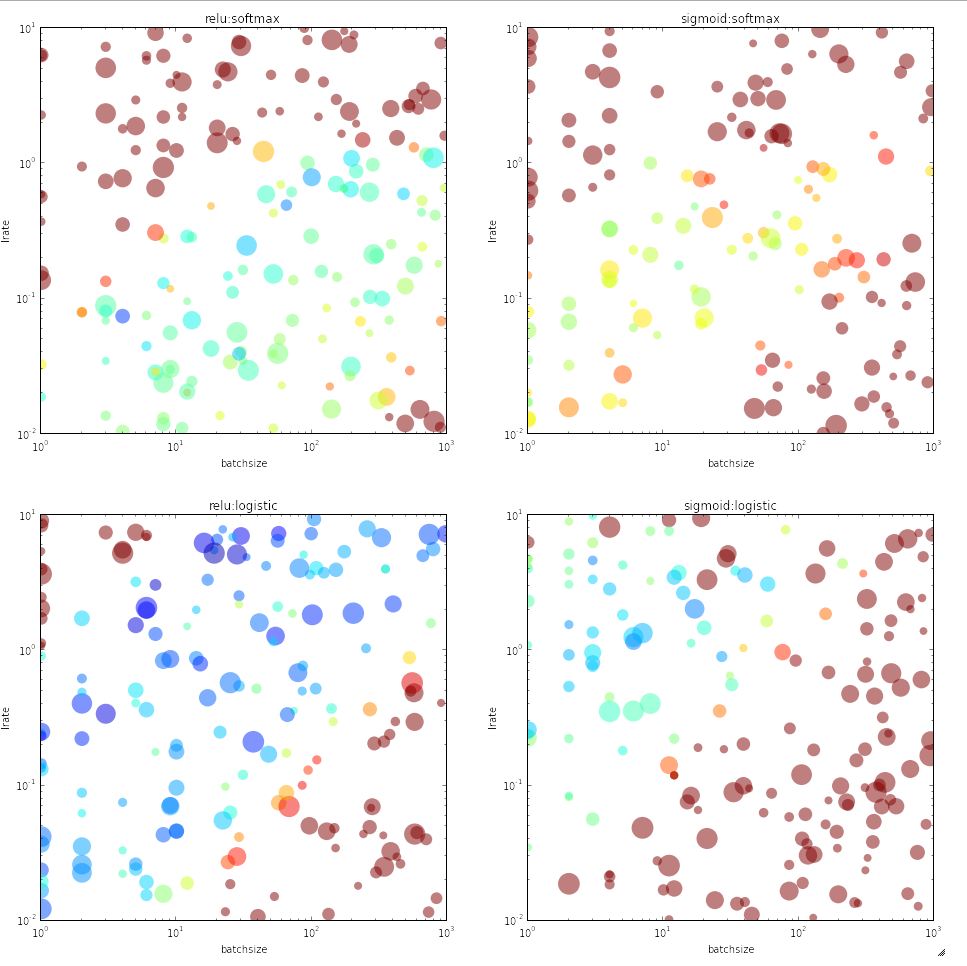}
\caption{\label{tsall}
Error rates (indicated by color) vs batch size, learning
rate, and number of hidden units (indicated by circle size). These
scatterplots suggest that the parameter ranges for all four conditions
were explored fairly completely.
}
\end{figure}

\section{ReLU Units (Convolutional)}

The previous results for ReLU units are interesting, given the low test
set error rate and low dependence on batch size. It’s interesting to
see whether we can reproduce those results for convolutional networks.
However, in the case of convolutional networks, we find a significant
batch size dependence, and that the difference between sigmoidal and
ReLU hidden units is considerably smaller than for non-convolutional
networks. Furthermore, some softmax networks also come close in
performance (Figure 13).

\begin{figure}[tp]
\includegraphics[width=3in]{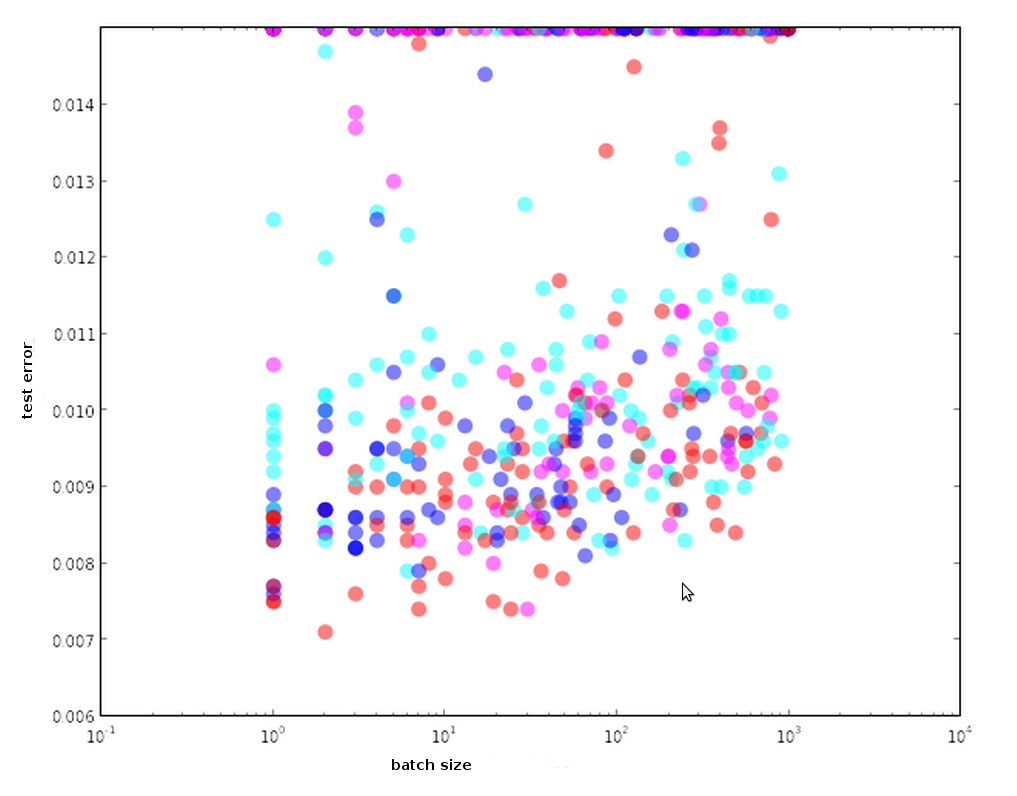}
\caption{\label{bsconvrelu}
Test error vs. batch size for convolutional networks with ReLU units.
}
\end{figure}

\begin{figure}[tp]
\includegraphics[width=4in]{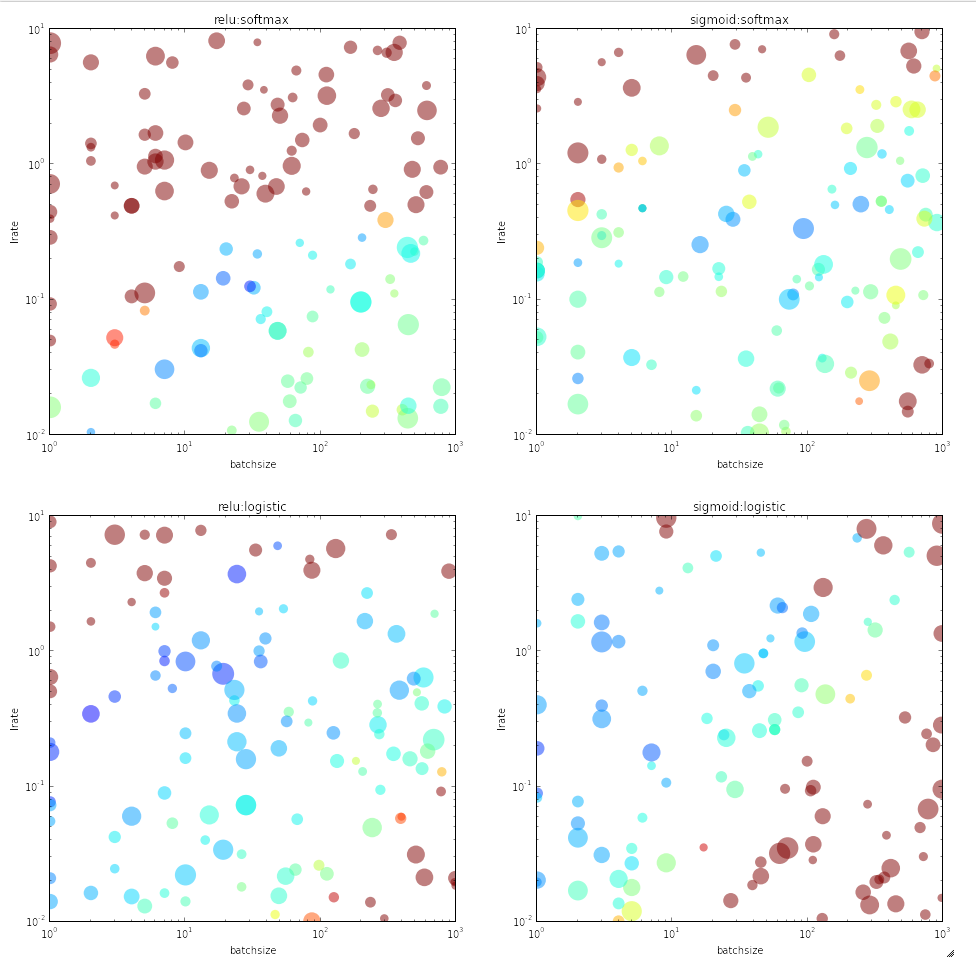}
\caption{\label{convparm}
The batch size and learning rate parameter space explored for
the comparison of convolutional ReLU and sigmoidal networks.
}
\end{figure}

\section{Deep ReLU Networks}

 ReLU networks also appear to be highly successful for training deep
networks. We explore this architectural variant in these experiments by
comparing deep networks with ReLU hidden units and sigmoidal hidden
units. We find that the performance of deep networks with sigmoidal
units degrades with depth, while deep networks with ReLU networks yield
good performance even at large (eight layer) depths (Figure 15).
However, increasing depth for ReLU networks does not result in better
test set performance. The deterioration of test set error with
increasing depth for sigmoidal hidden units is probably due to effects
like vanishing gradients, something that ReLU networks do not seem to
suffer from to the same degree.

\begin{figure}[tp]
\includegraphics[width=\textwidth]{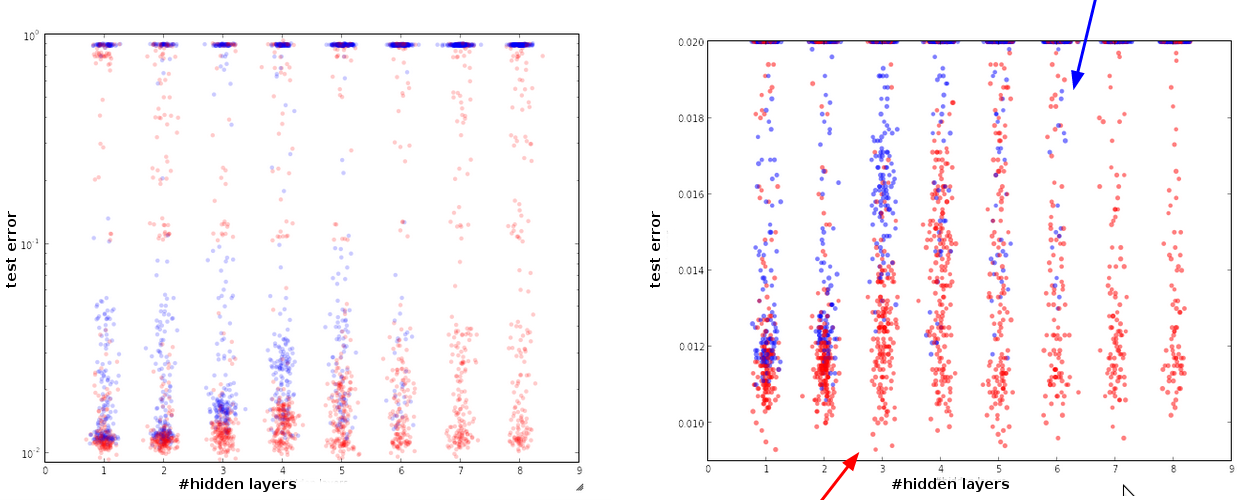}
\caption{\label{deepneterr}
 Test set error by network depth for networks with sigmoidal
hidden units (blue) and ReLU hidden units (red).
}
\end{figure}

For both types of deep networks, we can ask again how test set error
depends on batch size. Figure 16 shows that for both types of deep
networks and across a range of hidden layers, the best achievable test
set error generally increases with increasing batch size.

\begin{figure}[tp]
\includegraphics[height=3in]{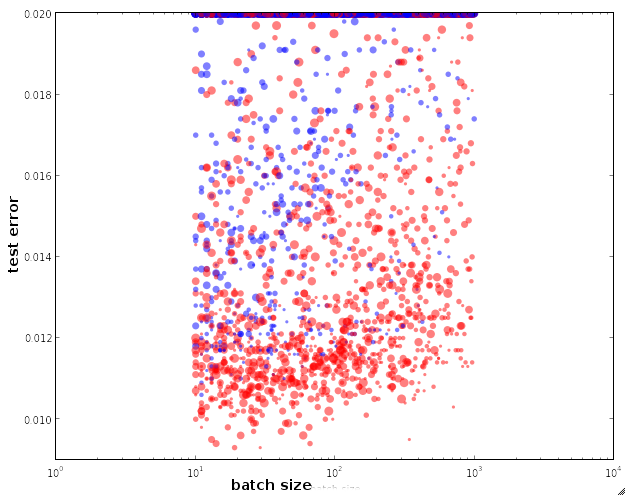}
\caption{\label{deepnetsize}
Test set error vs batch size for deep networks with ReLU
(red) and sigmoidal (blue) hidden layers. Circle size indicates number
of hidden layers. Notice that there is a significant decrease in test
set performance with increasing batch size.
}
\end{figure}

\begin{figure}[tp]
\includegraphics[width=\textwidth]{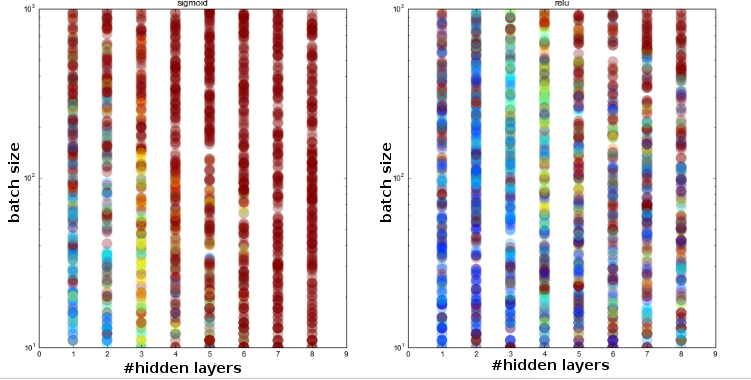}
\caption{\label{deepnetbs}
Test set error (indicated by color with red being high, blue
being low) vs \# hidden layers and batch size. Note that deep networks
are much more sensitive to large batch sizes during training than
shallow networks.
}
\end{figure}

\begin{figure}[tp]
\includegraphics[width=\textwidth]{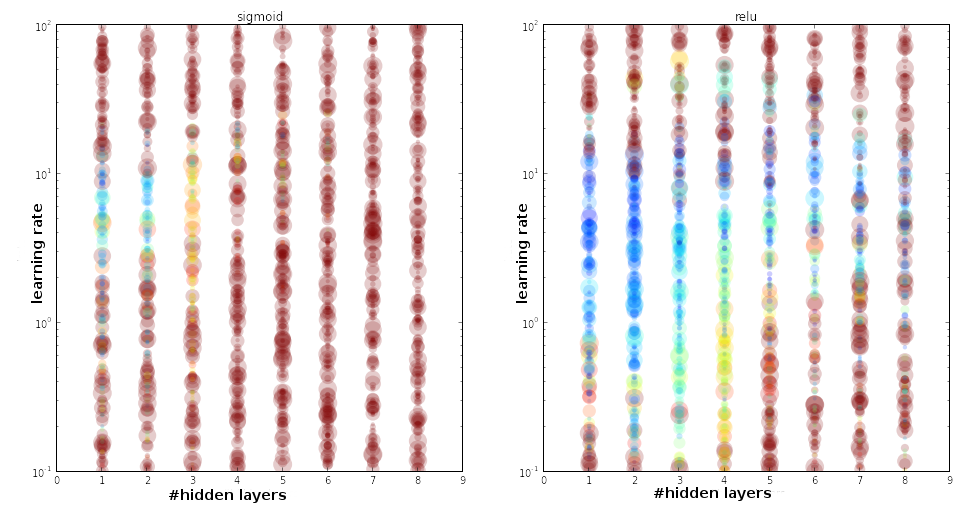}
\caption{\label{deepnetlr}
Parameters spaces explored by the deep ReLU and sigmoidal
networks. The parameter spaces show that the learning rate parameters
were explored sufficiently well for both network types.
}
\end{figure}

\section{Conclusions and Discussion}

Machine learning algorithms cannot be totally ordered by performance and
there is no single best learning algorithm.
Nevertheless, machine learning benchmarks in
general, and benchmarks on MNIST in particular, tell us something about
how machine learning algorithms compare on typical
classification tasks, and what kind of architectural features
influence performance significantly.
That is, the conclusions we can draw from such benchmarks
are not so much about which algorithm is “better”, but rather which
algorithmic choices may affect the outcome positively or negatively.

Perhaps the most important result from these benchmarks is how complex
the interaction between different architectural features and conditions
is; performance improvements that can be demonstrated in simple
networks do not add up when combined together into the same architecture.

Furthermore, many benchmarks that have been carried out in the literature
may have been hampered by limited sets of training conditions.
For example, benchmarking logistic vs. softmax outputs at larger batch
sizes suggests that there is little difference between the two methods;
however, at small batch sizes, logistic outputs significantly
outperform softmax outputs on MNIST data (Figure~\ref{bste}). Unless
both methods are tested at small batch sizes, the significantly better
performance of logistic outputs is not revealed.
As a second example, for some architectures, ReLU networks show no batch
size dependencies, while other network architectures do show such
dependencies.

It is important to remember that the MNIST dataset is not necessarily
representative of other classification problems: it has a small number
of classes, the prior probability is uniform, the number of training
samples is small compared to many other problems, all geometric
variability (translations, rotations, skew) has been removed, and the
input vectors are binary. Therefore, more important than the results
about what works are the result of what unexpectedly doesn't work
well even in such a simple case.

Based on the experiments reported here, we observe:

\begin{itemize}
\item For many problems, increasing batch sizes in a parallel
implementation results in no speedup in training because the per-sample
learning rate needs to be scaled down proportionately to batch size.
Furthermore, large batch sizes may intrinsically limit the performance
of networks. Finally, hyperparameter optimization may get harder for
larger batch sizes, as the range of feasible learning rates (and other
parameters) gets narrower. Therefore, it is a good idea to carry out
experiments with single sample updates and small batch sizes.
\item Softmax outputs may yield lower training errors than logistic
outputs, but often also yield higher test set errors. Therefore, it is
a good idea to try both kinds of outputs when training neural networks
on different tasks. In doing so, it is important to try a wide range of
learning rates, since the optimal learning rates for the two kinds of
outputs are different.
\item For non-convolutional networks, ReLU hidden layer units perform
significantly better than sigmoidal hidden layer units in these
experiments; they also show lower batch size dependencies and scale to
much larger numbers of hidden units. For convolutional networks,
however, these effects were not observed, suggesting that both
sigmoidal and ReLU non-linearities should be tried.
\item It is much easier to train deep networks using ReLU hidden layers
than hidden layers with sigmoidal non-linearities. However, additional
depth does not improve test set error for either sigmoidal or ReLU
units. For deep networks, we also observed batch size dependencies. In
addition, the range of good learning rates shifts and becomes smaller
for deeper networks.
\end{itemize}

\noindent
Generally, these results suggest the following strategy for training new
networks:

\begin{itemize}
\item Start with single sample updates, both during initial exploration
and hyperparameter search.
\item When exploring new problems, compare softmax and logistic outputs,
as well as ReLU and sigmoidal hidden units.
\item Although deep ReLU networks can be trained, be sure to try shallow
ReLU networks as well, and test a wide range of learning rates.
\end{itemize}

\noindent
We note that there has been an extensive literature on various
improved optimization methods for neural network learning, methods 
for learning hyperparameters, and benchmarks of MLP performance.
It is impossible to do this literature justice in this technical
report. However, a few general observations should suffice:

\begin{itemize}
\item The methods described in this paper all relied on simple SGD
    training, yet yield excellent performance compared to other
    reported results. In particular, optimization or hyperparameter
    selection methods that yield significantly worse results than
    those reported here are of questionable utility.
\item Generally speaking, hyperparameter optimization for these
    these kinds of problems does not seem to be particularly critical;
    networks yield similar performance over a broad range of 
    hyperparameters. 
\item Hyperparameter optimization should not optimize for the 
    best expected test set error of the resulting networks, but
    for the minimal error over a collection of multiple trained
    models.
\end{itemize}

\bibliography{mlpbench}
\bibliographystyle{plain}

\end{document}